\documentclass{article}

\usepackage{microtype}
\usepackage{graphicx}
\usepackage{subfigure}
\usepackage{booktabs} %

\PassOptionsToPackage{hyphens}{url} %

\usepackage{hyperref}

\usepackage{natbib}
\usepackage{graphicx}

\usepackage[accepted]{icml2024}

\usepackage{amsmath}
\usepackage{amssymb}
\usepackage{mathtools}
\usepackage{amsthm}

\usepackage[utf8]{inputenc} %
\usepackage[T1]{fontenc}    %
\usepackage{hyperref}       %
\usepackage{url}            %
\usepackage{booktabs}       %
\usepackage{amsfonts}       %
\usepackage{nicefrac}       %
\usepackage{color}
\usepackage{xcolor}

\usepackage{ dsfont }
\usepackage{xspace}
\usepackage{doi}
\usepackage[capitalize,noabbrev]{cleveref}

\usepackage{enumitem}
\setlist{noitemsep,topsep=-\parskip}
\theoremstyle{plain}
\newtheorem{theorem}{Theorem}[section]

\theoremstyle{definition}
\newtheorem{definition}[theorem]{Definition}

\theoremstyle{remark}

\newcommand{\tuple}[1]{\langle #1 \rangle}

\newcommand{\reals}{\mathbb{R}}
\newcommand{\MMDP}{multi-stakeholder MDP\xspace}

\newcommand{\init}{s_\text{init}}
\newcommand{\MMDPelements}{S,\init,\allowbreak A,P,R_1,\dots,\allowbreak R_n, \gamma}
\newcommand{\MMDPtuple}{\tuple{\MMDPelements}}

\DeclareMathOperator*{\expectedvalue}{\mathbb{E}}
\newcommand{\return}{G}

\newcommand{\defeq}{\stackrel{\text{def}}{=}}

\DeclareMathOperator*{\argmax}{arg\,max}

\newcommand{\donut}
{doughnut\xspace}
\newcommand{\donuts}{doughnuts\xspace}

\newcommand{\statusnamelong}{stakeholder status\xspace}
\newcommand{\statusname}{status\xspace}

\newcommand{\fairscheme}{\mathfrak{F}}

\newcommand{\aggname}{aggregation\xspace}

\newcommand{\trace}[2]{\tau_{#2}}%

\newcommand{\fairNMDP}{NMFDP\xspace}

\newcommand{\fairMDPtupleTiny}{\tuple{\mathcal{M}, \fairscheme}}

\newcommand{\fairMDPtupleExpanded}{\tuple{\MMDPtuple, \allowbreak\fairschemetuple}}
\newcommand{\fairMDPtupleSchemeExpanded}{\tuple{\mathcal{M}, \fairschemetuple}}
\newcommand{\fairMDPtupleExpandedBoundless}{\tuple{\MMDPtuple,\allowbreak\fairschemetupleboundless}}

\newcommand{\fairschemeelements}{\status,\tempagg,\filter}%
\newcommand{\fairschemetuple}{\tuple{\fairschemeelements}}
\newcommand{\fairschemetupleboundless}{\tuple{\status,\allowbreak\tempagg}}

\newcommand{\status}{U}
\newcommand{\agg}{W}
\DeclareMathOperator{\mean}{mean}

\newcommand{\hor}{T} %

\newcommand{\memory}{M}
\newcommand{\minit}{m_\text{init}}
\newcommand{\mupdate}{\mu}

\newcommand{\filter}{B}
\newcommand{\filtername}{filter\xspace}

\newcommand{\tempagg}{W_\text{ex}}
\newcommand{\tempaggname}{extended aggregation\xspace}
\newcommand{\timeagg}{W_\text{temporal}}
\newcommand{\timeaggname}{temporal aggregation\xspace}
\newcommand{\unfairness}{\text{unfairness}}

\newcommand{\timefirst}{timepoint-first\xspace}
\newcommand{\Timefirst}{Timepoint-first\xspace}

\newcommand{\util}{\mathit{Util}}
\newcommand{\rawls}{\mathit{Rawls}}
\newcommand{\nash}{\mathit{Nash}}

\usepackage[textsize=tiny]{todonotes}
\usepackage[normalem]{ulem}

\newcommand{\methodNameFull}{Fair Q-Learning with Counterfactual Memories\xspace}
\newcommand{\methodName}{FairQCM\xspace}

\icmltitlerunning{Remembering to Be Fair: Non-Markovian Fairness in Sequential Decision Making}

\newcommand{\documentStyle}{two-columns}

\ifthenelse{\equal{\documentStyle}{one-column}}{%
    \newcommand{\expPlotWidth}{0.8\columnwidth} %
}{%
    \newcommand{\expPlotWidth}{\columnwidth} %
}

\begin{document}

\twocolumn[
\icmltitle{Remembering to Be Fair: \\Non-Markovian Fairness in Sequential Decision Making}

\icmlsetsymbol{equal}{*}

\begin{icmlauthorlist}
\icmlauthor{Parand A. Alamdari}{equal,uoft,vectorinst,sri}
\icmlauthor{Toryn Q. Klassen}{equal,uoft,vectorinst,sri}
\icmlauthor{Elliot Creager}{waterloo,vectorinst,sri}
\icmlauthor{Sheila A. McIlraith}{uoft,vectorinst,sri}
\end{icmlauthorlist}

\icmlaffiliation{uoft}{University of Toronto, Toronto, Canada}
\icmlaffiliation{waterloo}{University of Waterloo, Waterloo, Canada}
\icmlaffiliation{vectorinst}{Vector Institute, Toronto, Canada}
\icmlaffiliation{sri}{Schwartz Reisman Institute for Technology and Society, Toronto, Canada}

\icmlcorrespondingauthor{Parand A. Alamdari}{parand@cs.toronto.edu}

\icmlkeywords{Fairness, Machine Learning}

\vskip 0.3in
]

\printAffiliationsAndNotice{\icmlEqualContribution} %

\begin{abstract}
Fair decision making has largely been studied with respect to a single decision. Here we investigate the notion of fairness in the context of sequential decision making where multiple stakeholders can be affected by the outcomes of decisions. We observe that fairness often depends on the history of the sequential decision-making process, and in this sense that it is inherently non-Markovian. We further observe that fairness often needs to be assessed at time points \emph{within} the process, not just at the end of the process. To advance our understanding of this class of fairness problems, we explore the notion of non-Markovian fairness in the context of sequential decision making. We identify properties of non-Markovian fairness, including notions of long-term, anytime, periodic, and bounded fairness. 
We explore the interplay between non-Markovian fairness and memory and how memory can support construction of fair policies. Finally, we introduce the FairQCM algorithm, which can automatically augment its training data to improve sample efficiency in the synthesis of fair policies via reinforcement learning.
\end{abstract}

\section{Introduction}
\label{sec:intro}

\begin{figure}[t]
    \centering
\subfigure{
\includegraphics[width=\expPlotWidth,trim=3cm 3.5cm 11cm 3cm, clip]{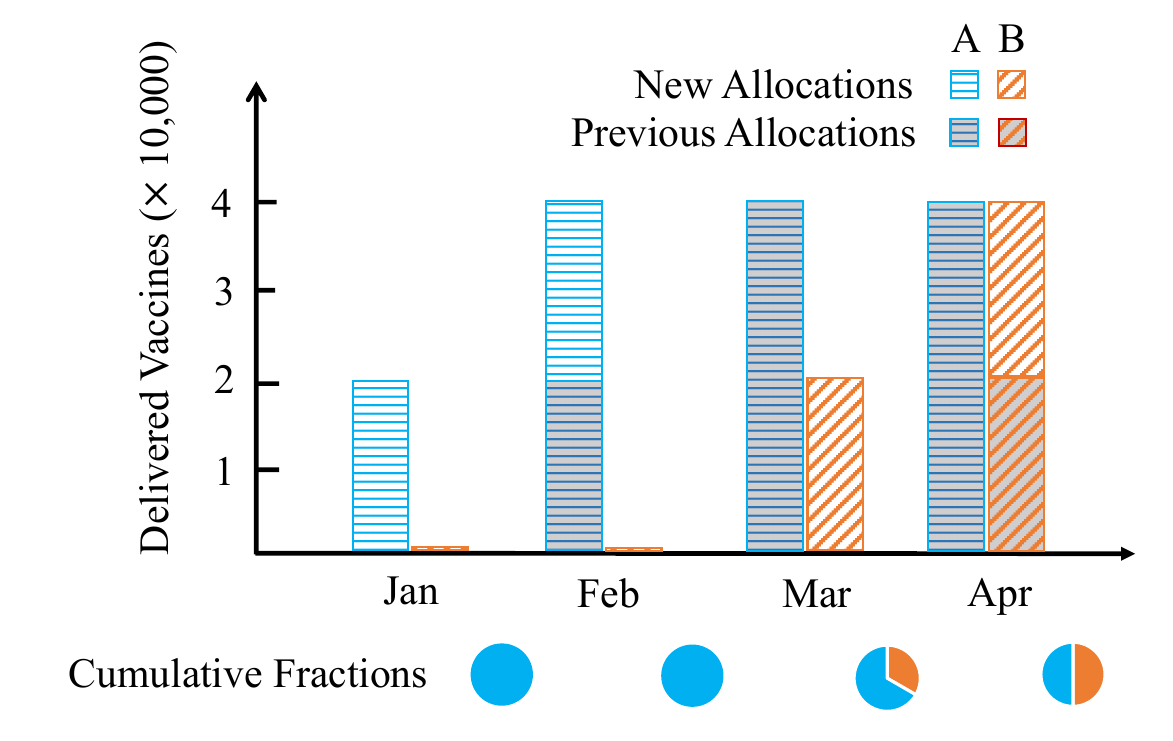}
}
\subfigure
{
\includegraphics[width=\expPlotWidth, trim=3cm 3.5cm 11cm 3cm, clip]{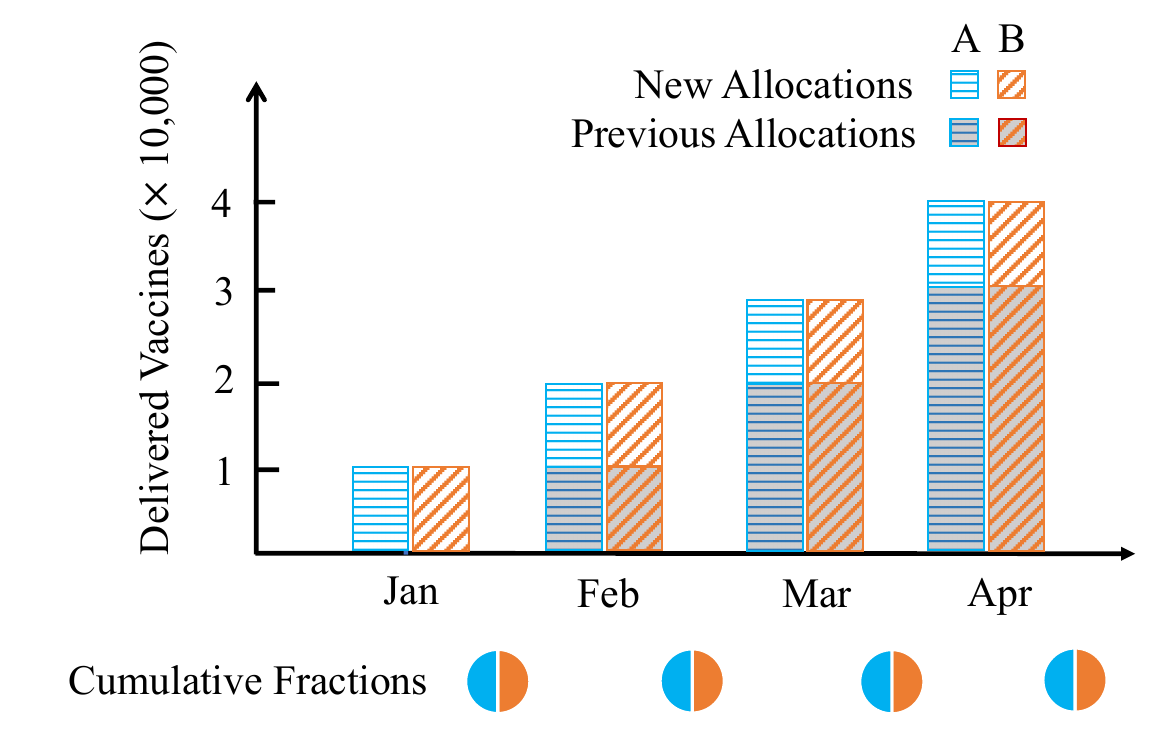}
}

\caption{Two processes for distributing vaccines to countries A and B. Both result in an equal distribution of vaccine at the end. Monthly evaluation shows that the first %
process favors A for a time.}
    \label{figs:vaccine}
\end{figure}

In many real-world decision-making settings, decisions are of consequence to a diversity of entities---the \emph{stakeholders} of the decision-making process.
Furthermore, decision making is often sequential, involving multiple decisions executed over time in support of realizing near- and/or longer-term objectives. 
Here we study 
fairness in the context of sequential decision making, where multiple stakeholders can be affected by the outcomes of decisions.

To ground this discussion, consider the problem of distributing vaccines to countries around the world, and our aspiration that the allocation to different countries be “fair” in some manner. 
Much of the decision making required to distribute vaccines  
is in service of getting vaccine safely from origin to destination. Delivery can be complex and costly, restricting the set of feasible plans. Indeed many such decisions will not immediately affect the fairness of the vaccine distribution. Does it even make sense to ask whether these intermediate decisions are “fair?” Perhaps what we need to aspire to is \emph{long-term fairness} of the entire decision-making process, in the limit when all the vaccines have been delivered. Unfortunately, this has its problems.

To illustrate, consider two greatly simplified processes to distribute 80,000 vaccines to countries A and B, as depicted in \Cref{figs:vaccine}; the first process reflects an economical plan that delivers 40,000 vaccines to country A in two consecutive monthly shipments of 20,000 each, followed by two consecutive monthly shipments of the same size to country B.  In the second process, A and B each receive 10,000 vaccines each month for four months. Each process results in the delivery of 40,000 vaccines 
 to each of A and B. However the latter process may seem ``more fair,''
particularly when we treat early vaccine access as a proxy for 
reduced negative health outcomes \citep[e.g.,][]{DuBMC2024covid}.

This simple example suggests the need to measure and enforce fairness at significant timepoints throughout the process---a notion of \emph{periodic} fairness, perhaps monthly, quarterly, or at year's end. In the extreme we might aspire to a notion of \emph{anytime fairness}, 
or perhaps a notion of \emph{bounded fairness} where the fairness of the allocation of vaccines is judged at a time point that is determined by some property of the system, such as after delivering each million vaccines.

This example also exposes an important property of fairness in sequential decision making, that it can be inherently non-Markovian. 
That is, the assessment of fairness of a sequential decision-making process does not just rely on the current state, or more generally $(s,a,s')$, the state $s'$ resulting from deciding to perform action $a$ in state $s$. Rather, it is a function of the history of the decision-making process, the history of state-action pairs. This has implications not only for how we define various notions of fairness %
but also how we compute fair policies, and the role of memory.

To the best of our knowledge, this is the first work to explore these important concepts. Our contributions include:

\begin{enumerate}[topsep=0pt,itemsep=-1ex,partopsep=1ex,parsep=1ex]
\item We introduce the notion that multi-stakeholder fairness can be non-Markovian in sequential decision making and define formative concepts relating to the assessment of fairness at varying time intervals including long-term, periodic, anytime, and bounded fairness. 
\item We study the role of memory in converting non-Markovian problems into Markovian problems 
so that the generation of fair policies, which are inherently non-Markovian, can be addressed using standard Markovian as well as non-Markovian methods.
\item We propose \methodName, an algorithm that encourages sample-efficient fair policy learning by generating counterfactual experiences during the learning process. We demonstrate that \methodName promotes sample-efficient fair learning compared with baseline methods that learn explicitly from the entire history, or use a neural memory module. 
\end{enumerate}
These contributions are relevant both to the synthesis of fair sequential decision making, and to the assessment of fairness with respect to historical traces of human- and/or machine-generated decision making.

There are many practical instances of sequential decision making problems where the assessment of fairness is potentially non-Markovian in nature and where some form of intermittent fairness assessment and enforcement may be necessary.
There is a rich literature on vaccine allocation \cite{Erdogan2024pandemic}.
Interestingly, the COVID vaccine allocation program COVAX had the explicitly temporal objective of participating countries progressing \emph{at the same rate}, and they developed an allocation algorithm \cite{who2021allocation} whose inputs included the historical allocation of vaccines from the program.\footnote{We note that COVAX has been subject to various criticisms \citep[see, e.g.,][]{UsherLancet2021COVAX}.}
Another practical example of sequential decision making, where the assessment and optimization of fairness can require consideration of history, is in establishing fair \emph{waiting times} in healthcare, such as for hospital admissions or surgeries \citep[e.g.,][]{QiMS2017delays, AlaSR2021whale}.

We mention these applications to reinforce the practical import and potential impact of the framing and foundational ideas presented in this paper.  To simplify our discussion going forward, we will migrate from our vaccine example to an even simpler example involving the distribution of indivisible goods---in this case doughnuts.

\noindent {\bf Running Example -- Doughnut Allocation:  }Given $n$ people, and %
$m$ doughnuts for distribution, %
define a policy to distribute these doughnuts in a fair manner over time. This problem presents at least two challenges: (i) defining what constitutes a fair allocation of these goods, and (ii) prescribing, or \emph{learning} a policy to realize a fair distribution. We will use variants of this problem to illustrate our framework.

\section{Related Work}\label{sec:related}

We are not the first to study fairness over time.
Indeed, a flurry of research activity has followed from the observation that intervening to promote fairness in the short-term can lead to distinct and sometimes unexpected results in the long run~\citep{liu2018delayed,hashimoto2018fairness,hu2018short,d2020fairness}.

\citet{zhang2014fairness} introduced a method to maximize returns for the worst-off agent in a setting where agents have local interests.
\citet{jabbari2017fairness} explored a more conservative notion of fairness, where a reinforcement learning (RL) agent was tasked with considering long-term discounted rewards when comparing two actions, reminiscent of a dynamic take on the individual fairness principle of ``treating likes alike''~\citep{dwork2012fairness}.
Group notions of fairness have also been explored and theoretically analyzed within RL. \citet{wen2021algorithms} offer practical approaches to encourage parity in rewards between a majority and minority group, as well as characterizations of the sample complexity involved. \citet{deng2022reinforcement} consider group fairness constraints at each step and provide theoretical guarantees for fairness violations.

Defining what constitutes fairness is an open-ended design choice with normative and political implications~\citep{narayanan2018translation,binns2018what,xinying2023guide}. Given the importance of the scalar reward signal in RL, one key question is how to aggregate rewards across stakeholders and time to measure the overall fairness of a policy.

A (cardinal) social welfare function aggregates individual utilities together into one number.
Suppose we have a sequence of utilities $\vec u=u_1,\dots,u_n$. Some well-known social welfare functions are the utilitarian welfare $\util(\vec u)=\sum_i u_i$, the Rawlsian welfare $\rawls(\vec u)=\min (u_1,\dots, u_n)$, and the Nash welfare $\nash(\vec u)=\prod_i u_i$ \citep[see, e.g.,][]{CaragiannisTEAC2019nash}. Some of these can be viewed, in part, as measuring fairness.
Recent work has proposed optimizing Nash welfare, Gini social welfare, and utilitarian objectives~\citep{Mandal2022socially,SiddiqueICML2020fair,FanAAMAS2023welfare}.
Our contributions are distinct in our focus on measuring fairness over a history of decisions---as opposed to immediate choices or long-term aggregated discounted rewards---and the special role of memory that this non-Markovian perspective on fairness entails.

Our motivating sequential doughnut allocation example is related to an established literature on allocation problems~\citep{ibaraki1988resource}, where fairness concerns have also been explored~\citep{kash2014no,bouveret2016characterizing,CaragiannisTEAC2019nash}.
Our focus on formal frameworks for sequential decision making differentiates us in this regard; although temporally-extended notions of fairness have also been suggested in the context of computational social choice~\citep{BoehmerAAMAS2021broadening}, they are underexplored, even in this context.

\section{Preliminaries}
\label{sec:preliminaries}

We consider an environment that
is fully observable 
with dynamics that
are governed by a Markov Decision Process (MDP) \citep{puterman2014markov}. We utilize an MDP variant that makes explicit the stakeholders that are affected by the sequential decision-making process and their individual reward functions. Our definition is functionally equivalent to that of  %
a multi-objective MDP \citep{RoijersJAIR13multiobjective}.

\begin{definition}[Multi-stakeholder Markov Decision Process\xspace]
A \MMDP is a tuple $\MMDPtuple$ where
 $S$ is a finite set of states, $\init \in S$ is the initial state, $A$ is a finite set of actions, 
and $P(s_{t+1}\mid s_t, a_t)$ is the transition probability distribution, giving the probability of transitioning to state $s_{t+1}$ by taking action $a_t$ in $s_t$. 
For $i=1,\ldots,n$, $R_i:S\times A\times S\to \reals$ is the reward function of the $i$th stakeholder, and $\gamma \in(0,1]$ is the discount factor. 
\end{definition}
In such an environment, if executing action $a_t$ in state $s_t$ results in state $s_{t+1}$, each stakeholder $i$ receives the reward $R_i(s_t, a_t, s_{t+1})$. The formalism is agnostic with respect to who---stakeholders or others---actually executes the actions.

A \textit{(Markovian) policy} $\pi(a\mid s)$ is a probability distribution over the actions $a \in A$, given a state $s \in S$. We can also define a \emph{non-Markovian policy} as a mapping from histories to distributions over actions: $\pi(a_{t}\mid s_1,a_1,\dots, a_{t-1},s_{t})$.

A \emph{trace} $\tau$ of a \MMDP is a (possibly infinite) sequence of alternating states and actions:   $s_1,a_1,s_2,a_2,s_3,a_3,\dots$ (where $s_1=\init$). %
A finite trace always ends with a state.
Following in the spirit of \citet{sutton2018reinforcement}, given a trace $\tau$ we can define the \emph{discounted return} for stakeholder $i$ as the discounted sum of that stakeholder's rewards accumulated over the trace:
$\return_i(\tau)\defeq\textstyle\sum_{t} \gamma^{t-1} R_i(s_t, a_t, s_{t+1})$
($t$ ranges over the timepoints in the trace, whether there are finitely or infinitely many.)
A policy may generate various traces (i.e., when actions are selected by it) and could be evaluated in terms of \emph{expected} discounted returns.

For an MDP, which can be thought of as a \MMDP where $n=1$, it is standard to try to find a policy that maximizes the expected discounted return. 
It is well-known that (for infinite traces) there always exists a Markovian policy that does so
\citep{Watkins1992qlearning}.

Which policies are to be preferred when there are more stakeholders? Each stakeholder might prefer a policy that gives itself higher expected discounted return. We consider a number of ways to assess a policy's fairness in \Cref{sec:non-markovian-fairness}.

Finally, here is some further  notation we use:
\begin{itemize}

\item For a finite trace
$\trace{x}{\hor}=s_1, a_1, s_{2}, \dots, s_\hor, a_\hor, s_{\hor+1}$ we may use a subscript {\small $T$} to indicate the length (number of actions) as just shown. 
\item Given a trace $\tau=s_1, a_1, \dots, s_{t+1},a_{t+1},$ $s_{t+2},\dots$ of length $>t$, we can write $\trace{x}{t}$ for the prefix  $s_1,\allowbreak a_1,\allowbreak\dots, s_{t+1}$. Recall that a finite trace ends with a state.
\item For $n \in \mathbb{N}$, we define the set $[n] \defeq \{1,2,...,n\}.$
\end{itemize}

\section{Fairness over Time}
\label{sec:non-markovian-fairness}

In \Cref{sec:intro} we observed that the assessment of fairness depends on some function of the history of states and actions and consequently is non-Markovian. In this section we argue for the importance of \emph{non-Markovian} fairness and propose (\Cref{sec:schemes}) a formalism that allows for expressing many different forms of fairness over time.

\noindent {\bf Example:~}Returning to our simple \donut example, consider the case where we have 3 children (stakeholders) A, B, C, and 24 \donuts to distribute. There are three actions $giveA$, $giveB$, $giveC$ that, in the deterministic case, make the propositional variables $gotA$, $gotB$, and $gotC$ true, respectively. Not surprisingly received \donuts disappear near instantaneously! 
This is a simple allocation problem and there are many ways of characterizing fairness, but for ease of explication, a simple strategy to assess whether, at the end of the process, the 24 \donuts were allocated fairly is to inspect the history, count the number of times each child received a \donut and if the sums for each child agree, we can proclaim fairness of the process.  $\blacksquare$

To formalize this intuition,
suppose that we have a function $\status:(S\times A)^*\times S\to \reals^n$ that tells us how well each stakeholder is doing at the endpoint of the given trace. 
The idea is that $\status(\trace{1}{\hor})$'s $i$th entry, $\status(\trace{1}{\hor})_i$, is some measure of how good $\trace{1}{\hor}$ has been to stakeholder $i$. 
We will call $\status$ the
\emph{\statusnamelong function}, and the value it returns the \emph{\statusnamelong vector}.

In our \donut example, the stakeholder status function would be a function over the entire history, resulting in a vector of size $n=3$ denoting the number of \donuts distributed to each of A, B, and C.
For other applications, possible \statusnamelong functions could be stakeholder returns or some measure of how much \emph{envy} stakeholder $i$ feels on $\tau$ \citep[see][]{ShamsADT2021envy}.

Suppose also for now that we have some \emph{\aggname} function $\agg:\reals^n\to \reals$ that we can use to aggregate stakeholder scores to determine how fair things are. $\agg$ could for instance be a standard social welfare function like Nash welfare (\Cref{sec:related}), or a boolean-valued function which outputs 1 just in case all stakeholder scores are equal. This is indeed what is required for our \donut example.

Given $\status$ and $\agg$, when can we say that a trace $\tau$ was fair? It might seem that we have an obvious answer: to equate the fairness of the trace with fairness at the end.
\begin{definition}[Long-term fairness]
Given a \statusname function $\status$ and an \aggname function $\agg$, 
the \emph{long-term fairness} of a finite trace $\trace{1}{\hor}$ is $\agg(\status(\trace{1}{\hor}))$.
\end{definition}

As argued in \Cref{sec:intro}, in many cases, we wish to assess the fairness of a process more frequently, perhaps weekly or monthly, rather than just at the end. To do so, we define a notion of periodic fairness.
There is a variety of options for aggregating periodic assessments. One simple option would be to take the \statusname vector at the end of each week, apply the \aggname function, and then sum the results. In such a case, if comparing traces of equal length, the greater the sum, the more fair the trace. 
For now let us just say that we have some function $\tempagg:(\reals^n)^*\to\reals$ that lets us aggregate a sequence of \statusnamelong vectors. We will call $\tempagg$ the \emph{\tempaggname} function (higher scores are better).

\begin{definition}[Periodic fairness]
Given a \statusname function $\status$ and \tempaggname function $\tempagg$,
the \emph{periodic fairness} (with period $p$) of a finite trace $\trace{1}{\hor}$ is  
$\tempagg\left(\status(\trace{1}{p}),\status(\trace{1}{2p}),\dots,\status(\trace{1}{\lfloor \hor/p \rfloor p})\right)$.
\end{definition}
A special case is to assess the trace at every time point:
\begin{definition}[Anytime fairness]
\label{def:anytime}
Given a \statusname function $\status$ and \tempaggname function $\tempagg$,
the \emph{anytime fairness} of a trace $\trace{1}{\hor}$ is its periodic fairness with period 1.
\end{definition}
Long-term fairness could be thought of as a special case of anytime fairness -- the case in which the \tempaggname function $\tempagg$ ignores all but the last  of its inputs.

A generalization of periodic fairness is \emph{bounded fairness}, where we have some ``\filtername'' function $\filter(\trace{1}{t})$ to indicate at which points fairness should be assessed (for example, after a dozen doughnuts has been distributed):
\begin{definition}[Bounded fairness]
Given a \statusname function $\status$, an \tempaggname function $\tempagg$, and a binary-valued function $\filter(\trace{1}{t})$,
the \emph{bounded fairness} of a finite trace $\trace{1}{\hor}$ is 
$\tempagg\left(\status(\trace{1}{t_1}),\status(\trace{1}{t_2}),\dots,\status(\trace{1}{t_k})\right)$ where $(t_1, t_2, \dots, t_k)$ is the subsequence of $(1,2\dots,\hor)$ for which $\filter(\trace{1}{t_i})=1$ for each $t_i$.
\end{definition}

In \Cref{sec:schemes}, we bring these notions together into a formalism that can express many conceptions of fairness for traces and also for policies.

\subsection{Fairness Schemes}
\label{sec:schemes}

To compare traces (and later policies) in a general way, we introduce \emph{fairness schemes}, which include as elements the \statusname, \tempaggname, and \filtername functions.

\begin{definition}[Fairness scheme]\label{def:fairnessscheme}
Given a \MMDP with state space $S$ and action space $A$, a \emph{fairness scheme} is a tuple $\fairscheme=\fairschemetuple$ where\\
\noindent ~~\textbullet\ 
$\status: (S\times A)^*\times S \to \reals^n$ is the \emph{\statusnamelong function}.\\ 
\noindent ~~\textbullet\  
$\tempagg:(\reals^n)^*\to\reals$ is the \emph{\tempaggname function}.\\
\noindent ~~\textbullet\ 
$\filter:(S\times A)^*\times S\to\{0,1\}$ is the \emph{\filtername function}.\\
In some cases we may not specify a \filtername function and just write $\fairscheme=\fairschemetupleboundless$, which means that we assume that $\filter$ is the constant function $\filter(\tau)=1$ (indicating that fairness should be assessed at all time points.)
\end{definition}

We typically pair a fairness scheme with a \MMDP, so for convenience we have the following definition:
\begin{definition}[Non-Markovian Fair Decision Process (\fairNMDP)]
A \emph{Non-Markovian Fair Decision Process} is a tuple $\fairMDPtupleTiny$ where $\mathcal{M}$ is a \MMDP and $\fairscheme$ is a fairness scheme defined w.r.t. $\mathcal{M}$.  
\end{definition}
Now, we can define the fairness score of a trace.
\begin{definition}[Fairness score of a trace]\label{def:fairscoretrace}
    Given an \fairNMDP $\fairMDPtupleTiny$ where $\fairscheme=\fairschemetuple$, the \emph{fairness score} of a finite trace $\trace{1}{\hor}$, which with a slight abuse of notation we will write as $\fairscheme(\trace{1}{\hor})$, is
    $$\fairscheme(\trace{1}{\hor})\defeq
    \tempagg(\status(\trace{1}{t_1}),\status(\trace{1}{t_2}),\dots, \status(\trace{1}{t_k}))$$ where $(t_1, t_2, \dots, t_k)$ is the subsequence of $(1,2\dots,\hor)$ for which $\filter(\trace{1}{t_i})=1$ for each $t_i$.
\end{definition}
That is, the fairness score is the \tempaggname of the \statusname vectors that pass the filter (note that this definition coincides with bounded fairness).

In the \donut example, $t_i$ could, for example, denote each time at which another dozen \donuts have been distributed and $\status(\trace{1}{t_i})$ could be a vector corresponding to the number of \donuts distributed to each of A, B, and C at that time. $W_{ex}$ then corresponds to a function over that sequence of vectors that measures their fairness.  

In the case where we are considering policies rather than historical traces, we can similarly use a fairness scheme $\fairscheme$ to score a policy by taking an expectation:
\begin{definition}[Fairness score of a policy]
    Given an \fairNMDP $\fairMDPtupleTiny$ and a time horizon $\hor$, the \emph{fairness score of a policy} $\pi$ according to fairness scheme $\fairschemetuple$ is
    \begin{align*}
    &\fairscheme^{\hor}(\pi)\defeq \expectedvalue_{\trace{1}{\hor}\sim\pi,\mathcal{M}}[\fairscheme(\trace{1}{\hor})]
    \end{align*}
    where $\fairscheme(\trace{1}{\hor})$ is the fairness score of the trace per \Cref{def:fairscoretrace}.
To evaluate fairness over infinite traces, we can consider the limit (if it exists) as $\hor\to\infty$%
; i.e., $\fairscheme^{\infty}(\pi)\defeq \expectedvalue_{\tau\sim\pi,\mathcal{M}}\left[\lim_{\hor\to\infty}\fairscheme(\trace{1}{\hor})\right].$
\end{definition}

For a given choice of fairness scheme, an {\bf optimal policy} would be the one that maximizes the fairness score, either for some given time horizon or in the limit. 
Finding such a policy can be considered in different settings with or without access to the dynamics model (if we do have access, then a form of planning might be used; otherwise, we might use RL). Furthermore, we might or might not have access to the fairness scheme itself (perhaps fairness scores just come as feedback from the environment).
We will consider computing policies in \Cref{sec:computation}.

\subsection{More on the \tempaggname function}

For the \tempaggname function $\tempagg$, one option would be to define it to first aggregate stakeholders statuses at each time point, and then aggregate across time.

\begin{definition}[\Timefirst]\label{def:timefirst}
Given a fairness scheme $\fairschemetuple$, we will say the \tempaggname function is \emph{\timefirst} if it can be written as 
\begin{align*}
    \tempagg(u_1,\dots, u_k) = \timeagg(\agg(u_1),\dots,\agg(u_k))
\end{align*}
in terms of two aggregation functions: an \aggname function $\agg:\reals^n\to\reals $ that is applied to the \statusnamelong vector at each time point, and a \timeaggname function $\timeagg:\reals^*\to\reals$ that combines the results.
\end{definition}

Here are some example \timeaggname functions:\\
\noindent ~~\textbullet\ Long-term fairness: $\timeagg(w_1,\dots, w_k)=w_k$\\
\noindent ~~\textbullet\ Average: $\timeagg(w_1,..., w_k)=\mean (w_1,..., w_k)$\\ 
\noindent ~~\textbullet\ The discounted sum of the aggregated values for each time point (treating them like rewards): $$\timeagg(w_1,\dots, w_k) =\textstyle\sum_{t=1}^k \gamma^{t-1} w_t$$
Furthermore, $\timeagg$ could be a standard social welfare function, where we treat the aggregations from different time points as the individual utilities to be aggregated.
For example, we could be ``Rawlsian with respect to time'' and only care about the worst-off timepoint:
$\timeagg(w_1,\dots, w_k)=\min (w_1,\dots, w_k).$

However, not all desirable \tempaggname functions are naturally expressible in a \timefirst way. What if a trace is unfair \emph{to} a particular stakeholder at one time, and unfair \emph{in their favor} at another? We might want to say that those events balance out. However, from a single real number $\agg(\status(\trace{1}{t}))$, it is awkward to extract who is being (un)fairly treated at $t$. That information would be easier to define in terms of $\status(\trace{1}{t})$ itself. We consider this next.

\subsubsection{Unfairness for Individual Stakeholders}
\label{sec:unfairto}

We can use the \statusname function $\status$ to define when a specific stakeholder is being unfairly treated. In the \donut example where 24 \donuts were being distributed, a distribution of 8 \donuts to each of A, B, and C would constitute a fair distribution, whereas a distribution of 6, 8, and 10, respectively, would correspond to a distribution that was \emph{unfair to} A \emph{in favor of} C. 

\begin{definition}[Unfair to / unfair in favor of]
    Given an \fairNMDP $\fairMDPtupleTiny$ with \statusname function $\status$, a trace $\tau$  is \emph{unfair to} stakeholder $i$ at time $t$ on the trace if $\status(\trace{1}{t})_i < \mean(\status(\trace{1}{t}))$, and is \emph{unfair in favor of} stakeholder $i$ at time $t$ on the trace if $\status(\trace{1}{t})_i > \mean(\status(\trace{1}{t}))$.
\end{definition}

We can also make a quantified version of unfairness:
\begin{definition}[Unfairness to / overall unfairness to]
\label{def:unfairness}
    Given an \fairNMDP $\fairMDPtupleTiny$ with \statusname function $\status$, the \emph{unfairness} of trace $\tau$ to stakeholder $i$ at time $t$ on the trace is $\unfairness_i(\tau,t)\defeq \status(\trace{1}{t})_i - \mean(\status(\trace{1}{t})).$
    The \emph{overall unfairness} of a finite trace $\trace{1}{\hor}$ to $i$ is
    $
     \unfairness_i(\trace{1}{\hor})\defeq\textstyle\sum_{t\in[\hor]} \unfairness_i(\trace{1}{\hor},t).
     $
\end{definition}

An \tempaggname function could take unfairness to individuals into account in various ways; for example, we could set $\tempagg$ to be the negative of the sum of squared overall unfairness values: $-\textstyle\sum_{i\in[n]}(\unfairness_i(\trace{1}{\hor}))^2$.

\section{Computing Fair Policies}
\label{sec:computation}

In this section we consider approaches to computing policies for \fairNMDP{s} in certain cases---for fairness schemes that can be made to behave like reward functions.

\subsection{The Role of Memory}
\label{sec:memory}

How should actions be selected in a \fairNMDP? In general, an optimal policy may need to consider the whole trace $\tau$ that has occurred so far, even just to determine what the current \statusnamelong vector is.
In this section, we will focus on how to transform the \fairNMDP to make the \statusnamelong function Markovian, in the following sense:

\begin{definition}[Markovian function]\label{def:markovianfunction}
    Given an \fairNMDP $\fairMDPtupleExpanded$, we will say a function $f:(S\times A)^*\times S\to \text{codomain}(f)$ is Markovian if %
\begin{align*}
    f(s_1,a_1,\dots, s_t, a_{t+1}, s_{t+1}) = f_\memory( s_{t+1})
\end{align*}
    for some function $f_\memory$; i.e., $f$ only varies with the last state.%
\end{definition}

Having a Markovian \statusname function $\status$ does not guarantee the existence of an optimal policy that is Markovian---that still depends on the rest of the fairness scheme $\fairscheme=\fairschemetuple$. (Also, in the finite horizon case the policy might have to be non-stationary.) However, there exist choices of the \tempaggname function $\tempagg$ and \filtername function $\filter$ that ensure there will be optimal Markovian policies. 

Suppose $\filter(\tau)=1$  (which we will be assuming until \Cref{sec:computefilter}). Suppose also that $\tempagg$ is \timefirst (\Cref{def:timefirst}) and so we have that
$\tempagg(\status(\trace{1}{1}),\dots, \status(\trace{1}{\hor}))=\timeagg(\agg(\status(\trace{1}{1})),\allowbreak\dots, \agg(\status(\trace{1}{\hor})))$,
and further that $\timeagg$ treats its inputs like rewards (e.g., by returning their discounted sum). Then if $\status$ is Markovian we can apply standard MDP techniques -- treating the composite function $\agg\circ\status$ as a reward function -- to find a Markovian policy. Indeed, to do so we would only need the weaker condition that $\agg\circ \status$ is Markovian.

Therefore, it is of interest to consider how to make the \statusname function Markovian. In some cases this can be done by augmenting the state space $S$ to remember past information.\footnote{To augment states with some form of memory is an idea with a long history \citep[e.g.,][]{BacchusBG96,PeshkinICML1999memory,ThiebauxJAIR2006decision,icarte2020act}.}
\begin{definition}[Memory augmentation/augmented]
Given an \fairNMDP $\fairMDPtupleExpandedBoundless$, a \emph{memory augmentation} is a tuple $\tuple{\memory,\minit, \mupdate}$ where $\memory$ is the finite set of memory states, $\minit\in \memory$ is the initial memory state, and $\mupdate:\memory\times A\times S\to \memory$ is the memory update function.
The resulting \emph{memory-augmented} \fairNMDP is an \fairNMDP
$
    \tuple{\tuple{S\times M, \tuple{\init,\minit}, A, P', R_1',\dots, R_n', \gamma},\allowbreak\tuple{\status',\tempagg}}
$
where
\begin{itemize}[noitemsep]
\item $P'(\tuple{s_{t+1}, m_{t+1}}\mid\tuple{s_{t},m_t}, a_t) = P(s_{t+1}\mid s_{t}, a_t)$ if $m_{t+1} = \mupdate(m_{t}, a_t, s_{t+1})$ and otherwise 0.
\item $R'_i(\tuple{s_t,m_t},a_t,\tuple{s_{t+1},m_{t+1}})= R_i(s_t,a_t,s_{t+1})$ for each $i$
\item $\status'(\tuple{s_1,m_1}, a_1, \dots, \tuple{s_t,m_t},a_t,\tuple{s_{t+1},m_{t+1}} )=\status(s_1,a_1,\dots, s_t, a_{t+1}, s_{t+1})$
\end{itemize}
\end{definition}
The idea is that the memory starts storing the value $\minit$, and given a trace $s_1, a_1, \dots, s_t, a_t, s_{t+1},\dots$  the value of the memory at time $t+1$ would be $m_{t+1}=\mupdate(m_t,a_t,s_{t+1})$ where $m_t$ is the value of the memory at time $t$. The original environment's dynamics are preserved, meanwhile.

While we wrote $\status'$ as not depending on the memory, the cases we are interested in are those in which it is Markovian in the memory-augmented \fairNMDP, and so is equivalent to a function $\status'_\memory(\tuple{s_{t+1},m_{t+1}})$.
There is a whole class of \statusname functions, which we will call \emph{value-regular}, for which we will show that there is a memory augmentation that makes them Markovian.
\begin{definition}[value-regular]
    Given an \fairNMDP $\fairMDPtupleExpandedBoundless$, the \statusnamelong function $\status$ is \emph{value-regular} if there is some finite set $V\subseteq \reals^n$ such that 
    for any finite trace $\tau$ we have that $\status(\tau)\in V$, 
    and for each $v\in V$ the set of traces that $\status$ maps to $v$ is \emph{regular} in the following sense: the set of strings $\{\sigma\in(A\times S)^*:\status(\init, \sigma)=v\}$ is a regular language (e.g., could be described with a regular expression) \citep[see, e.g.,][]{Sipser1997theory}.
\end{definition}

\begin{theorem}\label{thm:regularfairness}
    Let $\fairMDPtupleExpandedBoundless$ be an \fairNMDP where $\status$ is value-regular.
    Then there exists a memory augmentation $\tuple{\memory,\minit, \mupdate}$ so that, in the resulting memory-augmented \fairNMDP, %
    $\status'$ is Markovian.
\end{theorem}
For the proof, see \Cref{sec:proofs}.
It turns out that value-regular \statusnamelong functions are the \emph{only} ones which we can make Markovian through memory augmentation, per the following theorem (again, the proof is in \Cref{sec:proofs}).
\begin{theorem}
\label{thm:inverse-regular}
Let $\fairMDPtupleExpandedBoundless$ be an \fairNMDP and $\tuple{\memory,\minit, \mupdate}$ be a memory augmentation such that, in the resulting memory-augmented \fairNMDP, $\status'$ is Markovian. Then $\status$ is value-regular.
\end{theorem}

Some obvious choices of the \statusnamelong function are not value-regular in general---in particular any function that can take an unbounded number of values, for instance where $\status$ gives the (possibly discounted) return for each stakeholder. In practice, however, policies may not be run long enough for memory limitations to come into play, and so we could still store the stakeholders' returns in memory.

\subsection{Counterfactual Memories for RL}
\label{sec:counterfactual}

In this section we consider applying RL to learn policies for (certain) memory-augmented \fairNMDP{s}, and present an RL algorithm that takes advantage of the memory structure.

The idea of a \emph{reward-like} fairness scheme  (defined below) is that we can treat a \fairNMDP with one like an MDP with a single (Markovian) reward function. 
\begin{definition}[Reward-like]\label{def:rewardlike}
Given an \fairNMDP $\fairMDPtupleExpandedBoundless$,
we will say that the fairness scheme $\fairschemetupleboundless$ is \emph{reward-like} if
\begin{itemize}
\item $\tempagg$ is \timefirst, so
$\tempagg(u_1,\dots, u_\hor) = \timeagg(\agg(u_1),\dots,\agg(u_\hor))$,
\item $\timeagg(w_1,\dots, w_\hor )=\textstyle\sum_{t=1}^\hor \gamma^{t-1}w_t$,
\item and $\agg\circ\status$ is Markovian (in the sense of \Cref{def:markovianfunction}) (this will always be the case if $\status$ is Markovian).
\end{itemize}
\end{definition}

If we define the (Markovian) reward function as 
$R(s_t,a_t,s_{t+1})=R(s_{t+1})=\agg(\status(s_{t+1}))$, 
the objective of a policy is to maximize the expected discounted sum of rewards (here, we will assume an infinite time horizon).
For a \fairNMDP with a reward-like fairness scheme, we can therefore apply standard RL algorithms. When the \fairNMDP is memory-augmented, we can also do a bit more.

A standard reinforcement learner in an MDP collects experiences of the form $(s_t,a_t,s_{t+1}, R(s_t,a_t,s_{t+1}))$, where $s_t$ is the state at time $t$ and $R$ is the reward function, with which to learn a policy.
In a memory-augmented \fairNMDP where the fairness scheme is reward-like, the learner can do the same, and we can write an experience as
\setlength{\abovedisplayskip}{3pt}
\setlength{\belowdisplayskip}{3pt}
\begin{equation}
\big(\tuple{s_t,m_t},a_t,\tuple{s_{t+1},m_{t+1}},R(\tuple{s_{t+1},m_{t+1}})\big)\label{eq:realexperience}
\end{equation}
where $\tuple{s_t,m_t}$ is the memory-augmented state at time $t$.
However,
whenever the learner has an experience as in \Cref{eq:realexperience},
we can use our knowledge of the memory augmentation $\tuple{\memory,\minit, \mupdate}$ (and of $R$) to generate additional experiences of the form \begin{equation}\big(\tuple{s_t,m_t'},a_t,\tuple{s_{t+1},m'_{t+1}},R(\tuple{s_{t+1},m'_{t+1}})\big)\label{eq:counterfactualexperience}\end{equation}
where $m_t'\in\memory$ is any memory state and $m'_{t+1}=\mupdate(m_t',a_t,s_{t+1})$.
That is, we generate the experience the learner would have had, had it been in the counterfactual memory state $m_t'$ instead of $m_t$. Generating this requires only the ability to pick $m_t'\in\memory$ and query $\mupdate$ and $R$.

Such counterfactual experiences can then be used in training, at least for \emph{off-policy} RL methods that do not have to learn from experience collected using the current policy. For instance, tabular Q-learning \citep{Watkins1992qlearning} could be modified to, after each real experience in the environment, generate (some subset of) the possible counterfactual experiences as in \Cref{eq:counterfactualexperience}, and use all of the experiences to update the Q-function. We call the resulting algorithm 
\textbf{\methodNameFull (\methodName)}
(see \Cref{alg:qcm}).
For methods like DQN \cite{mnih2015human} that use a replay buffer, counterfactual experiences could be added to the buffer. Action selection, similar to classic Q-learning, can work with a variety of choices. A commonly adopted strategy is epsilon-greedy. 
\begin{algorithm}[h]
   \caption{Tabular \methodName}
   \label{alg:qcm}
\begin{algorithmic}
   \STATE {\bfseries Input:} A memory-augmented \fairNMDP $\tuple{\tuple{S\times M,\allowbreak\tuple{\init,\minit},A, P', R_1',\dots, R_n', \gamma},\tuple{\status',\tempagg}}$ with reward-like fairness scheme and $\gamma<1$, and learning rates $(\alpha_1,\alpha_2,\dots)$.
   \STATE Let $R=\agg\circ\status'$
   \STATE Initialize the $Q$-function 
   \STATE Let $\tuple{s_1,m_1}=\tuple{\init,\minit}$
       \FOR{each time $t=1,2,3,\dots$}
            \STATE Select some subset $\memory'_t\subseteq \memory\setminus\{m_t\}$. 
            \STATE Select some action $a_t$, and execute $a_t$ in the current state $\tuple{s_t,m_t}$ to acquire the experience\\$\big(\tuple{s_t,m_t},a_t,\tuple{s_{t+1},m_{t+1}},R(\tuple{s_{t+1},m_{t+1}})\big)$
            \FOR{each $m_{t}'\in\memory'_t$}
                \STATE Construct the counterfactual experience\\$\big(\tuple{s_t,m_t'},a_t,\tuple{s_{t+1},m'_{t+1}},R(\tuple{s_{t+1},m'_{t+1}})\big)$
                where $m'_{t+1}=\mupdate(m_t',a_t,s_{t+1})$.
            \ENDFOR
            \FOR{each (real or counterfactual) experience $\tuple{x_t,a_t,\allowbreak x_{t+1}, r_{t+1}}$}
                \STATE\textcolor{gray}{/* a Q-learning update with learning rate $\alpha_t$ */}\\
                $Q(x_t,a_t)\xleftarrow{\alpha_t} r_{t+1}+\gamma \max_{a\in A} Q(x_{t+1},a)$
            \ENDFOR
        \ENDFOR
\end{algorithmic}
\end{algorithm}

This approach of generating counterfactual experiences is adapted from the CRM algorithm from \citet{ToroIcarteJAIR2022machines}. There, instead of memory states, they were dealing with states of a ``reward machine'', an automaton used to define a non-Markovian reward function. 

For tabular \methodName, we prove that (under some conditions) given a memory-augmented \fairNMDP with reward-like fairness scheme, it converges to the optimal $Q$-function, yielding the optimal policy\footnote{The optimal $Q$-function is a function $Q^*(x,a)$ which gives the expected discounted return in state $x$ if action $a$ is taken and afterwards the best possible action is taken in all future time steps. For an optimal policy $\pi^*$, $\pi^*(a\mid x)=0$ if $a\notin\argmax_{a'} Q^*(x,a')$.} 
(proof in \Cref{sec:proofs}). %
\begin{theorem}[Convergence of tabular \methodName]\label{thm:counterfactualconvergence}
    Let $n^i(x,a)$ be the time at which action $a$ is executed in state $x$ for the $i$th time (including ``executions'' in counterfactual experiences) in a run of \methodName.  If the selection of actions and counterfactual experiences and of the learning rates $(\alpha_1,\alpha_2,\dots)$ are such as to satisfy the conditions 
    from \citet[p. 282]{Watkins1992qlearning} that
    \begin{align*}
        0\le \alpha_n < 1,\quad \sum_{i=1}^\infty \alpha_{n^i(x,a)}=\infty,\quad\sum_{i=1}^\infty [\alpha_{n^i(x,a)}]^2<\infty
    \end{align*}
    for all $x$ and $a$, %
    then the $Q$-function computed by \methodName will converge (as $t\to\infty$) to the optimal with probability 1.
    
\end{theorem}

\subsection{Computations with Filter Functions}
\label{sec:computefilter}

Recall that, in general, a fairness scheme (\Cref{def:fairnessscheme}) includes a \filtername function $\filter$ that selects which points should be assessed for fairness.
We can generalize the notion of a \emph{reward-like} fairness scheme (\Cref{def:rewardlike}) to allow for non-trivial \filtername functions.
\begin{definition}[Bounded reward-like]\label{def:rewardlikebound}
Given an \fairNMDP $\fairMDPtupleSchemeExpanded$,
we will call the fairness scheme $\fairschemetuple$ \emph{bounded reward-like} if
\begin{itemize}
\item $\tempagg$ is \timefirst, so
$\tempagg(u_1,\dots, u_k) = \timeagg(\agg(u_1),\dots,\agg(u_k))$,
\item $\timeagg(w_1,\dots, w_k )=\textstyle\sum_{t=1}^k w_t$,
\item $\agg\circ\status$ is Markovian (this will always be the case if $\status$ is Markovian),
\item and $\filter$ is also Markovian. 
\end{itemize}
\end{definition}
Importantly, $\filter$ could be Markovian as a result of memory augmentation, in much the same way as we considered making $\status$ Markovian in \Cref{sec:memory}. For example, consider periodic fairness with period $p$, which corresponds to having $B(s_1,a_1,\dots, a_k,s_k)=1$ iff $k\equiv 0\pmod p$. That could be made Markovian by having a mod $p$ counter as memory.

In contrast to \Cref{def:rewardlike}, we assume a finite horizon, and set $\gamma = 1$. Therefore, there is no discounting in the sum computed by $\timeagg$.
For a bounded reward-like fairness scheme, it can be seen that the fairness score of a finite trace is the same as its undiscounted return using the following reward function $R$, allowing for standard algorithms for finite horizon MDPs:
\begin{align*}
    R(s_t,a_t,s_{t+1})=R(s_{t+1})=\agg(\status(s_{t+1}))\cdot \filter(s_{t+1})
\end{align*}

\section{Experiments}\label{sec:experiments}

In this section we compare different methods for designing the augmented memory via two simulation studies. 
Experiments show how generating counterfactual memories during RL can improve the overall fairness and sample efficiency of training in dynamic settings with multiple stakeholders. In each experiment, let $\fairMDPtupleTiny$ be an \fairNMDP where $\mathcal{M}$ is a \MMDP and $\fairscheme = \fairschemetuple$ is a fairness scheme defined w.r.t. $\mathcal{M}$, where $\tempagg$ is \timefirst, and $B(\tau) = 1$. 
Given the NMFDP, we implement \methodName and several baseline memory-augmented agents within a Deep RL framework. %
Analogous tabular results are in \Cref{sec:tabular}, a continuous experiment is in \Cref{app:extra-exp}, and
technical details of experiments are in \Cref{sec:experimental-details}.

\noindent {\bf Baselines.}
To evaluate how memory affects the ability to learn a fair policy, we evaluate several baselines based on Deep Q Networks (DQN)~\citep{mnih2015human} with different types of augmented memory units $M$:
\emph{Full} stores the entire stakeholder status $U(\tau_t)$ while 
\emph{Min} stores $U(\tau_t) - \min_i U(\tau_t)_i$; and
\emph{RNN} does not have a separate memory, instead it has an extra layer of GRU \citep{cho2014properties} to remember the past.
\emph{FairQCM} is a deep, DQN-based, version of our proposed method from \Cref{sec:counterfactual}, which we use with different types of memory.%

\subsection{Resource Allocation}
\label{exp:donuts}

We modeled our running example of doughnut allocation as a stochastic environment, featuring a doughnut shop that bakes one doughnut in each step. There are $n=5$ customers, with customer $i$ being in front of the counter with probability $p_i$ in each step. The decision to allocate the freshly baked doughnut is made by the server, who selects one customer for the allocation. If the chosen customer is not at the counter, the doughnut goes to waste. A state encapsulates the presence of individuals at the counter, and there are $n$ associated actions, each corresponding to the allocation of a doughnut to a specific customer. At step $t$, $\status(\tau_t)_i$ is the number of doughnuts allocated to customer $i$ so far. We use Nash welfare to define the aggregation function for each time point as $\agg(\status(\tau_t)) = \log (Nash(\status(\tau_t) + 1)) = \sum_{i \in [n]}\log(\status(\tau_t)_i + 1)$. 
The server's goal is to maximize the discounted sum of $\agg \circ \status$, treating it as the reward (except that on steps where the server wastes a doughnut, the reward is considered to be 0). The episode length is 100.  %

In this setting FairQCM stores $\status(\tau_t)$ in the memory (like the \emph{Full} baseline) and generates counterfactual experiences based on it. For this environment we have some extra memory-augmented baselines. \emph{Reset} stores $U(\tau_t)$ but if in a time $t$ all $U(\tau_t)_i$ are equal, it sets them to zero and counts from there for the next steps.
\emph{Oracle} and \emph{Random} are hard-coded solutions, with \emph{Oracle} realizing the optimal turn taking algorithm, and \emph{Random} taking random actions.%

\Cref{figs:donut-dqn} illustrates the accumulated Nash welfare scores at the end of the episode in different phases of training across five approaches with different types of augmented memory. FairQCM outperforms other algorithms. The RNN baseline learns an approximation of the history and achieves higher accumulated Nash welfare scores compared to some of the other baselines. The RNN also demonstrates faster adaptation to avoid wasting doughnuts, as it 
focuses on the recent past rather than the entire history.

\subsection{Simulated Lending}
\label{sec:lending}
\begin{figure}[t]
\begin{center}
\centerline{\includegraphics[ width=\expPlotWidth]{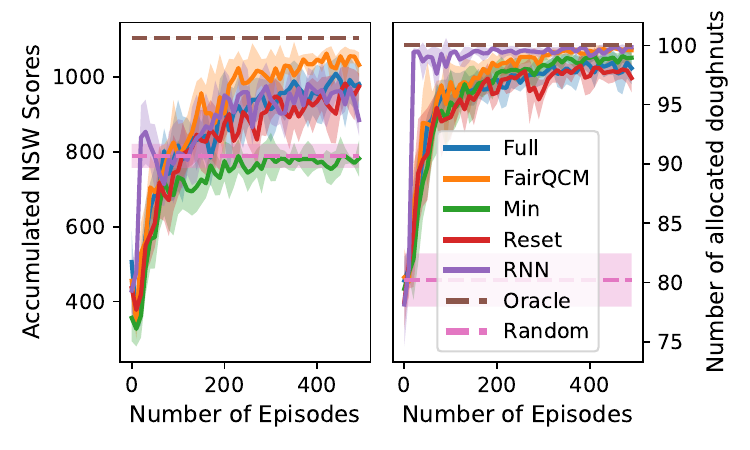}}
\caption{\textbf{Resource Allocation}: 
In simulations of our doughnut allocation task, (deep) \methodName achieves higher Nash welfare than competing memory-augmented RL agents (left), while learning to allocate doughnuts effectively near the end of training (right).}

\label{figs:donut-dqn}
\end{center}
\vspace{-2.5\baselineskip}
\end{figure}

Consumer lending is an established test bed for fair machine learning algorithms~\citep{dwork2012fairness,hardt2016equality}, owing in part to US regulations on credit scoring and banking practices~\citep{barocas2016big}.
As we are interested in fairness over time, we adapt the dynamic lending environment of \citet{liu2018delayed}. 
We consider a finite pool of loan applicants. Each applicant is characterized by their credit score $C$, representing the probability of loan repayment, and belongs to a protected group within the population (two groups in total: $A$ and $B$). The credit score distribution differs between the two groups.  In each step, a subset of applicants applies for a loan, and the bank must decide which applicant to grant the loan. Each applicant $i$ applies for a loan in each step with a probability of $p_i$. Successful loan repayment increases the bank's utility by $r$ and the applicant's credit score by $c$, while defaulting decreases the bank's utility by $r$ and the applicant's credit score by $c$. In our experiments, we set $r = 1$ and $c = 0.1$. The states encode the subset of applicants applying for a loan, their credit score, and profit margin so far. To assess the fairness of the loan granting process, we calculate the difference in the number of loans allocated to each protected subgroup. At step $t$, $\status(\tau_t)_i$ is the number of loans allocated to person $i$ so far. 
The aggregation function for each time point, which is inspired by demographic parity~\citep{dwork2012fairness}, we call the \emph{Relaxed Demographic Parity (DP) Score}
and is set to  {$\agg(\status(\tau_t)) = -|\sum_{i\in A} \status(\tau_t)_i - \sum_{i\in B} \status(\tau_t)_i|$. To ensure fairness, the bank's goal is to maximize the discounted sum of $\agg \circ \status$, treating that as the reward function -- with a couple exceptions. The bank aims to achieve a profit margin of at least 10 percent. If at the end of the episode it does not make the targeted profit, it incurs a substantial negative reward; furthermore, there is a negative reward for granting a loan to someone who didn't apply.
In the simulated lending environment, %
for the \emph{Full} baseline in this experiment, instead of storing each stakeholder status, we store $\status(\tau_t)_X = \sum_{i\in X} \status(\tau_t)_i$ in the memory for each protected subgroup $X \in \{A,B\}$. Similarly, the \emph{Min} baseline's memory stores $\tuple{\status(\tau_t)_A,\status(\tau_t)_B}-\min_{i \in \{A, B\}} \status(\tau_t)_i$. We use FairQCM with the same memory as the \emph{Min} baseline.

\Cref{figs:lending-dqn} illustrates the accumulated relaxed DP scores at the end of the episode in different phases of training across four approaches with different types of augmented memory. Similar to resource allocation, %
FairQCM outperforms the other approaches. The RNN baseline still learns an approximation of history through 40 steps of an episode and achieves higher accumulated relaxed DP scores compared to some of the other baselines. All approaches learn policies that achieve the desired profit margin of 10 percent.

The outcomes from experiments in distinct environments emphasize the complexity of selecting and engineering an appropriate memory solution. Algorithmically, FairQCM emerges as a standout performer with superior results, while RNN-based methods provide viable solutions, particularly where engineering a memory proves to be challenging. %

\begin{figure}[t]
\begin{center}
\centerline{\includegraphics[ width=\expPlotWidth]{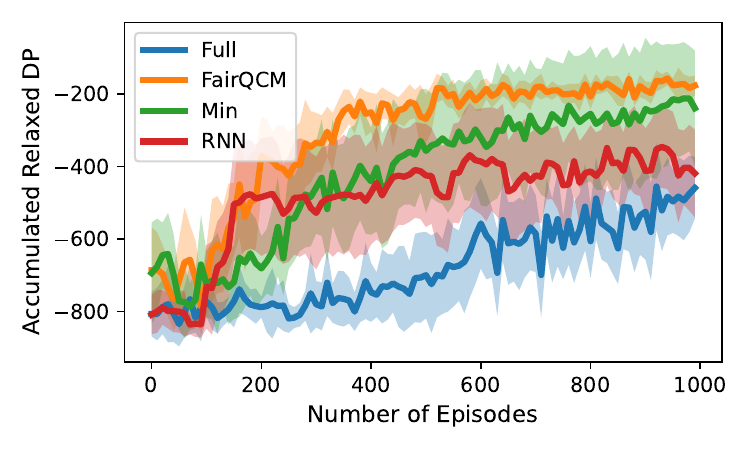}}
\caption{\textbf{Simulated Lending}: Accumulated Relaxed Demographic Parity scores for different approaches of augmenting memory during different phases of training.}
\label{figs:lending-dqn}
\end{center}
\vspace{-2.5\baselineskip}
\end{figure}

\section{Conclusion}
We have explored the notion of multi-stakeholder fairness in the context of sequential decision making. We have argued that the fairness of sequential decision making is inherently non-Markovian since the assessment of fair traces relies on the history of states and actions. We have also argued that fairness of processes is naturally assessed at significant time points, defining %
long-term, periodic, anytime, and bounded fairness. We have observed that in a number of circumstances, memory can be used to convert a non-Markovian fairness problem into a Markovian one, making it amenable to standard Markovian solutions---useful in the context of policy generation. We have studied the performance of various methods (Markovian and non-Markovian) and various memory models.
We also proposed a method called \methodName to generate counterfactual experience that expedite learning a fair policy, proved convergence properties in the tabular case, and demonstrated its effectiveness. %

We hope the contributions of this paper will provide the rich foundations for future work on fairness in sequential decision making---both in the area of policy generation and in auditing. A limitation of this work is that we have only scratched the surface in terms of exploring different fairness scoring functions and how to compute policies for them, and much work is left to do to formulate and assess our regime in practical settings in collaboration with domain experts.

\section*{Impact Statement}
Automated sequential decision making systems can affect many stakeholders, and thus have the potential to induce a variety of complex societal impacts.
In this paper we focus on how the fairness and overall social welfare of these systems evolves over time.
While our approach emphasizes the importance of historical decisions on near- and long-term fairness, we ultimately propose a flexible framework that is compatible with many existing fairness criteria.
For the proposed research to be socially beneficial, our methodological approach underscores the importance of a trained domain expert who can act in good faith to design appropriate fairness criteria that reflect normative commitments appropriate to the problem at hand.

\section*{Acknowledgements}
We wish to acknowledge funding from the Natural Sciences and
Engineering Research Council of Canada (NSERC), the Canada CIFAR AI Chairs
Program (Vector Institute), and Microsoft Research. The second author also received funding from Open Philanthropy.
Resources used in preparing this research
were provided, in part, by the Province of Ontario, the Government of Canada through CIFAR, and companies sponsoring the Vector Institute for Artificial Intelligence.

\bibliographystyle{icml2024-no-links}
\bibliography{ref}

\newpage
\appendix
\onecolumn
\section{Appendix}

\subsection{Proofs}
\label{sec:proofs}
\begin{proof}[Proof of \Cref{thm:regularfairness}]
        Let us say that the set of possible output values of $\status$ is $V=\{v_1,\dots, v_k\}$.
    Since $\status$ is value-regular, for each value $v_i\in V$ there exists a deterministic finite automaton (DFA) \citep[see, e.g.,][Chapter 1]{Sipser1997theory}
    $$A_{\status}^i =\tuple{Q^i,\Sigma,\delta^i,q_0^i, \mathit{Acc}^i}$$ 
    that accepts the string $\sigma=a_1,\dots, s_{t+1}$ iff $\status(\init,\sigma)=v_i$. Note that $Q^i$ is the finite set of automaton states, $\Sigma= A\times S$ is the set of input symbols, $\delta^i: Q^i\times\Sigma \to Q^i$ is the transition function, $q_0^i$ is the automaton's initial state, and $\mathit{Acc}^i\subseteq Q^i$ is the automaton's set of accepting states.

    We define a memory augmentation  $\tuple{\memory,\minit, \mupdate}$ as follows:
    \begin{align*}\memory &= Q^1\times Q^2\times \cdots \times Q^k\\
    \minit &= \tuple{q_0^1, q_0^2, \dots, q_0^k}\\
    \mupdate(\tuple{q^1,\dots, q^k},a,s)&=\tuple{\delta^1(q^1,\tuple{a,s}),\dots, \delta^k(q^k,\tuple{a,s})}
    \end{align*}
    That is, the memory keeps track of the states of the automata that correspond to each possible output value $v_1,\dots, v_k$. Then it can be seen that $\status'$ is Markovian in the memory-augmented \fairNMDP, because we can define
    \begin{align*}
        \status'_\memory(\tuple{s, \tuple{q^1,\dots, q^k}})=v_i\text{ iff } q^i\in \mathit{Acc}^i
    \end{align*}
    This is well-defined since the languages of the different automata are necessarily a partition of $(A\times S)^*$, so exactly one automaton will always be in an accepting state.
\end{proof}

\begin{proof}[Proof of \Cref{thm:inverse-regular}]
    $\status$ can only take the same values as $\status'$, and
since $\status'$ is Markovian, it can only take finitely many values (one for each augmented state $\tuple{s,m}$). Let us say that $V=\{v_1,\dots, v_k\}$ is the set of all possible output values of $\status$. For each $v_i$, we can construct a DFA $A_\status^i=\tuple{Q^i,\Sigma,\delta^i,q^i_0, \mathit{Acc}^i}$ where
\begin{itemize}
    \item $Q^i = S\times \memory$ is the set of automaton states,
    \item $q^i_0 = \tuple{\init,\minit}$ is the initial state,
    \item $\Sigma=A\times S$ is the alphabet,
    \item $\delta^i(\tuple{s,m},\tuple{a,s'})=\tuple{s', \mupdate(m,a,s')}$ is the transition function, and
    \item $\mathit{Acc}^i=\{\tuple{s,m}\in S\times \memory: \status'_\memory(\tuple{s,m})=v_i \}$ is the set of accepting states.
\end{itemize}
It can be shown by induction that the DFA $A_{\status}^i$ accepts the sequence $\sigma= \tuple{a_1, s_2}, \tuple{a_2, s_3}\dots, \tuple{a_t, s_{t+1}}$  if and only if $\status(\init,\sigma) = v_i$. It follows from the equivalence of DFAs and regular expressions that $\status$ is value-regular.
\end{proof}

\begin{proof}[Proof of \Cref{thm:counterfactualconvergence}]
The proof of convergence of Q-learning by \citet{Watkins1992qlearning} mostly carries over; 
the only thing we have to be careful about is whether the generated counterfactual experiences are biased relative to the environment's transition probabilities.\footnote{For example, if counterfactual experiences were only generated at time $t$ if the action $a_t$ resulted in a lottery being won, that could make winning the lottery seem more likely than it really is.} However, since the set $M_t'$ of counterfactual memory states to use in constructing the counterfactual experiences is chosen \emph{before} observing the outcome of action $a_t$, that issue does not arise: all action outcomes used in training are sampled according to the transition probabilities.
\end{proof}

\subsection{Tabular Q-Learning Experiments}
\label{sec:tabular}

We carry out the experiments described in \Cref{exp:donuts} in a tabular setting. The environment setting is similar to \Cref{exp:donuts}, where there are $n=3$ people in the doughnut shop in the tabular version, and the episode length is 12. We evaluate several baselines based on Q-Learning with different types of augmented memory units $M$: 

\begin{itemize}
    \item \emph{Full} stores the entire stakeholder status $U(\tau_t)$
    \item 
    \emph{FairQCM} stores the entire stakeholder status $U(\tau_t)$ as the memory $m_t$ at time $t$ and uses \Cref{alg:qcm} to generate counterfactual experiences. We limit it to 8 counterfactual experiences per time step, each such experience corresponding to a counterfactual memory $m_t'$ where for each $i$, $(m_t)_i < (m_t')_i \leq (m_t)_i + 2$. 
    \item \emph{Min} stores $U(\tau_t) - \min_i U(\tau_t)_i$
    \item \emph{Reset} stores $U(\tau_t)$ but if in a time $t$ all $U(\tau_t)_i$ are equal, it sets them to zero and counts from there for the next steps
    \item \emph{Oracle} and \emph{Random} are hard-coded solutions, with \emph{Oracle} realizing the optimal turn taking algorithm, and \emph{Random} taking random actions at each step.  
\end{itemize}

\begin{figure}[h!]
\begin{center}
    \begin{subfigure}
        \centering
        \includegraphics[width=0.45\textwidth]{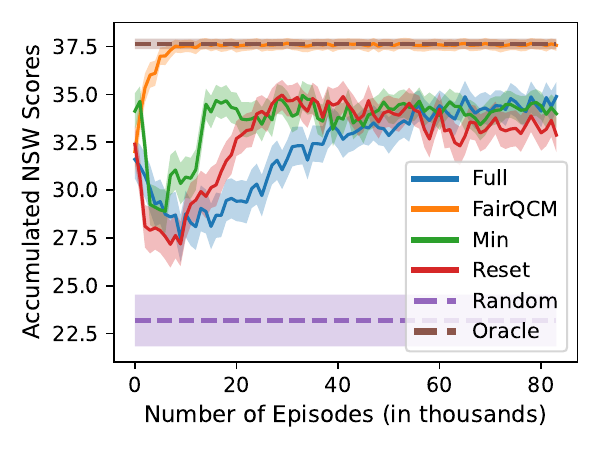}
        \label{fig:sub1}
    \end{subfigure}%
    \begin{subfigure}
        \centering
        \includegraphics[width=0.45\textwidth]{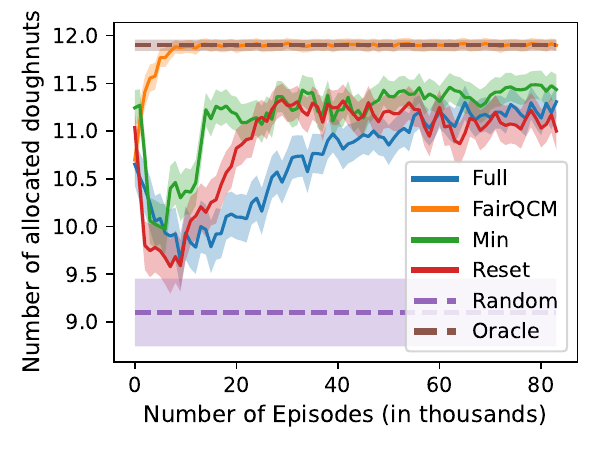}
        \label{fig:sub2}
    \end{subfigure}
\caption{\textbf{Tabular Q-Learning for Resource Allocation}: The left plot shows accumulated Nash welfare scores at the end of the episode for different approaches of augmented memory in different phases of training. The right plot shows the number of doughnuts that are not wasted during the process for each approach.}
\label{figs:donut-ql}
\end{center}
\end{figure}

Similar to the results in \Cref{sec:experiments}, FairQCM outperforms other approaches in terms of both NSW scores, and number of samples needed to train. 

\paragraph{Technical details} We set $\gamma = 0.99$, $\alpha = 0.1$, and use epsilon-greedy for exploration. $\epsilon = 1.0$ at the beginning for each state, and every time we visit a state $s$, $\epsilon_s$ is multiplied by $0.95$ (diminishing factor), and remains greater than $0.2$. We ran each method 10 times, and \Cref{figs:donut-ql} shows the average and variance of those runs across 100000 time steps.

\subsection{Extra Experiments with Simulated Lending Environment}

\label{app:extra-exp}
We extend the experiments described in \Cref{sec:lending}  in the simulated lending domain, to a scenario where the credit scores of applicants are continuous variables. The change in the credit scores (whether positive or negative) is sampled from a Gaussian distribution ($\mu = 0.05, \sigma = 0.1$ when the applicant successfully repays their loan, and $\mu = -0.05, \sigma = 0.1$ when the applicant defaults). 
The rest of the environment and the baselines are the same as \Cref{sec:lending}. \Cref{figs:lending-dynamic} illustrates the accumulated relaxed DP scores at
the end of the episode in different phases of training across
three approaches with different types of augmented memory. We observe the analogous behavior that FairQCM was more sample efficient, yielding superior fairness measures for a given number of episodes.

\begin{figure}[b]
\begin{center}
\centerline{\includegraphics[ width=0.5\expPlotWidth]{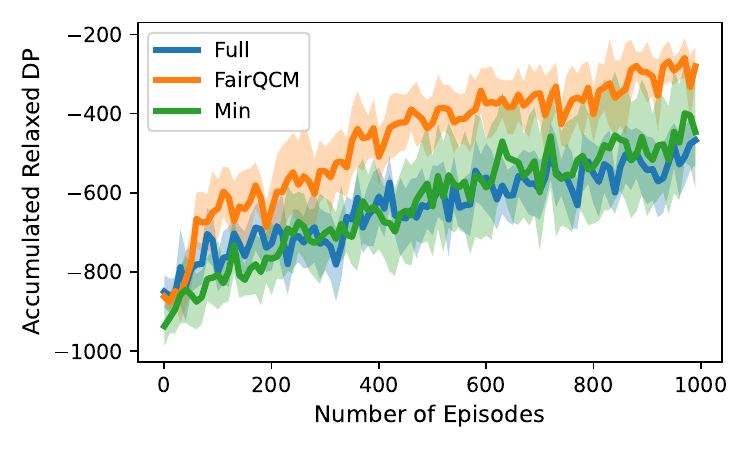}}
\caption{\textbf{Simulated Lending with Gaussian Credit Score Changes}: 
Accumulated Relaxed Demographic Parity scores for different approaches of augmenting memory during different phases of training.}

\label{figs:lending-dynamic}
\end{center}
\vspace{-\baselineskip}
\end{figure}

\subsection{Experimental Details}
\label{sec:experimental-details}
We ran the experiments on a system with the following specification: 2.3 GHz Quad-Core Intel Core i7 and 32 GB of RAM. The total running time is less than 24 hours. The code for all experiments is available at \href{https://github.com/praal/remembering-to-be-fair}{https://github.com/praal/remembering-to-be-fair}. 

\subsubsection{Resource Allocation}
\paragraph{State Representation} We consider $n = 5$ people at the doughnut shop. Each person is at the counter with $p = 0.8$. The state is a binary sequence of length $n$ showing the people present at the counter. 
\paragraph{Actions} There are $n$ actions which represent allocating a doughnut to person $i$ for each $i$. 

\paragraph{Rewards} $\status(\tau_t)_i$ represents the number of doughnuts person $i$ got so far. At time step $t$, if the doughnut is not wasted $r_t = \agg(\status(\tau_t)) = \log (Nash(\status(\tau_t) + 1)) = \sum_{i \in [n]}\log(\status(\tau_t)_i + 1)$ otherwise, $r_t = 0$.

\paragraph{Neural Network Architectures}
Our DQN consists of 4 fully connected layers with ReLU activation function: $\text{states} \times 32$, $32 \times 16$, $16 \times 8$, $8 \times \text{actions}$

RNN approach consists of 3 fully connected layers and one GRU layer: $\text{states} \times 32$, $32 \times 16$, $\text{GRU (hidden states = 256)}$, $16 \times \text{actions}$
\paragraph{FairQCM}
FairQCM stores the entire stakeholder status $U(\tau_t)$ as the memory $m_t$ at time $t$ and uses \Cref{alg:qcm} to generate counterfactual experiences. We limit it to 32 counterfactual experiences per time step, each such experience corresponding to a counterfactual memory $m_t'$ where for each $i$, $(m_t)_i < (m_t')_i \leq (m_t)_i + 2$. FairQCM stores the counterfactual experiences in the replay buffer, and later samples from them during training.

\paragraph{Results} We ran each method 10 times, and plot the average and variance of results across 1000 episodes in \Cref{figs:donut-dqn}.

\paragraph{Hyperparameters}
See \Cref{tab:dqn_hyperparameters}. 

\begin{table}[b]
\centering
\begin{tabular}{ |c|c|c|c| } 

        \hline
        \textbf{Hyperparameter} & \textbf{Full, Min, and Reset approaches}  & \textbf{RNN approach} & \textbf{FairQCM}\\
        \hline
        Episode Length & $100$ & $100$ & $100$\\
        \hline
        Learning Rate & $0.0001$ & $0.002$ & $0.0001$\\
        \hline
        Discount Factor ($\gamma$) & $0.95$ & $0.95$ & $0.95$\\
        \hline
        Min Exploration Rate ($\epsilon$) & $0.2$ & $0.2$ & $0.2$\\
        \hline
        Replay Buffer Size & $400$  & $1000$ & $6400$ \\
        \hline
        Batch Size & $64$ & $256$ & $2048$ \\
        \hline

    \end{tabular}
    \caption{Resource Allocation Hyperparameters}
    \label{tab:dqn_hyperparameters}
\end{table}
\subsubsection{Simulated Lending}

\paragraph{State Representation} We consider $n = 4$ people in total, where each group has two people in it. The initial credit score of people in group $A$ is $0.5$, and $0.9$ for group $B$. Each applicant applies for a loan at each step with $p = 0.9$.  The state consists of a binary sequence of length $n$ showing the applicants applying for a loan, credit score of each applicant, and profit margin so far. If an applicant gets a loan and repays it, their credit score increases by $0.1$, and similarly it decreases by $0.1$ if the applicant defaults. However, a credit score of an applicant always remains in the range of $[0.2, 0.9]$. 

\paragraph{Actions} There are $n$ actions which represent granting the loan to person $i$ for each $i$.

\paragraph{Rewards} The rewards in this encoding of the problem are serving two purposes. One purpose is to incentivize the system to learn to do the right thing---in this case, to only give loans to people who applied, and to maintain the stipulated profit margin.  In service of this, (i) at the final time step, if the bank didn't make the profit margin of 10 percent then $r_t = -10 \times \text{episode length}$; and (ii) if the bank grants a loan to someone who didn't apply for a loan then $r_t = -\text{episode length}$, where episode length is 40.

The reward is also used to reflect the fairness score.  
$\status(\tau_t)_i$ represents the number of loans person $i$ has received so far. At time step $t$, if the bank grants a loan to an applicant that applied for the loan then $r_t = \agg(\status(\tau_t)) = -|\sum_{i\in A} \status(\tau_t)_i - \sum_{i\in B} \status(\tau_t)_i|$.

\paragraph{Neural Network Architectures}
Our DQN consists of 3 fully connected layers with ReLU activation function: $\text{states} \times 32$, $32 \times 8$, $8 \times \text{actions}$.

RNN approach consists of 3 fully connected layers and one GRU layer: $\text{states} \times 32$, $32 \times 16$, $\text{GRU (hidden states = 256)}$, $16 \times \text{actions}$.

\paragraph{FairQCM} 
For each protected subgroup $X \in \{A, B\}$, FairQCM stores $\sum_{j \in X}U(\tau_t)_j - \min_{k \in \{A, B\}} \sum_{j \in k}U(\tau_t)_j$ as the memory $m_t$ at time $t$ and uses \Cref{alg:qcm} to generate counterfactual experiences. As each $m_t$ can be written as $m_t = \tuple{0, x}$ or $m_t = \tuple{x, 0}$ ($x \in [n]$), we limit the number of counterfactual experiences to at most 10 per time step, each such experience corresponding to a counterfactual memory $m_t'$ where, if $m_t = \tuple{0, x}$, $m_t' \in \{ \tuple{0, x-5},  \tuple{0, x-4}, \dots, \tuple{0, x-1}, \tuple{0, x+1}, \dots \tuple{0, x+5}\}$, and similarly if $m_t = \tuple{x, 0}$. FairQCM stores the counterfactual experiences in the replay buffer, and later samples from them during training.

\paragraph{Results} We ran each method 10 times, and plotted the average and variance of results across 1000 episodes in \Cref{figs:lending-dqn}.

\paragraph{Hyperparameters}
See \Cref{tab:dqn_hyperparameters-lending}.
\begin{table}[H]
\centering
\begin{tabular}{ |c|c|c|c| } 
        \hline
        \textbf{Hyperparameter} & \textbf{Full and Min approaches}  & \textbf{RNN approach} & \textbf{FairQCM}\\
        \hline
        Episode Length & $40$ & $40$ & $40$\\
        \hline
        Learning Rate & $0.0001$ & $0.005$ & $0.0001$\\
        \hline
        Discount Factor ($\gamma$) & $0.95$ & $0.95$ & $0.95$\\
        \hline
        Min Exploration Rate ($\epsilon$) & $0.2$ & $0.2$ & $0.2$\\
        \hline
        Replay Buffer Size & $1000$ & $2000$  & $8000$\\
        \hline
        Batch Size & $64$ & $512$ & $512$\\
        \hline
    \end{tabular}
        \caption{Simulated Lending Hyperparameters}
    \label{tab:dqn_hyperparameters-lending}
\end{table}

\end{document}